\documentclass[letterpaper, 10 pt, conference]{ieeeconf}  

\IEEEoverridecommandlockouts                              

\usepackage{graphicx} 
\usepackage{mathtools}

\usepackage{amsmath} 
\DeclareMathOperator*{\argmax}{argmax}

\usepackage{textcomp}
\usepackage{tabularx,booktabs}
\usepackage{color}
\usepackage[pdftex,dvipsnames]{xcolor}
\usepackage{array}
\usepackage{algorithm} 
\usepackage{algpseudocode} 
\usepackage{subcaption} 

\newcolumntype{P}[1]{>{\centering\arraybackslash}p{#1}}
\newcolumntype{C}{>{\centering\arraybackslash}X} 

\usepackage{tikz}

\title{\LARGE \bf
Multi-Technique Sequential Information Consistency For Dynamic Visual Place Recognition In Changing Environments
}

\author{Bruno Arcanjo$^{1}$, Bruno Ferrarini$^{1}$, Michael Milford$^{2}$, Klaus D. McDonald-Maier$^{1}$ and Shoaib Ehsan$^{1, 3}$
\thanks{}
\thanks{$^{1}$B. Arcanjo, B. Ferrarini, K. D. McDonald-Maier and S. Ehsan are with the School of Computer Science and Electronic Engineering, University of Essex, United Kingdom {\tt\small (email: bq17319@essex.ac.uk; bferra@essex.ac.uk; kdm@essex.ac.uk; sehsan@essex.ac.uk)}}%
\thanks{$^{2}$M. Milford is with the School of Electrical Engineering and Computer Science, Queensland University of Technology, Brisbane, QLD 4000, Australia
        {\tt\small (email: michael.milford@qut.edu.au)}}%
\thanks{$^{3}$S. Ehsan is also with the school of Electronics and Computer Science, University of Southampton, United Kingdom \tt\small{(email: s.ehsan@soton.ac.uk)}}%
}

\begin{document}

\maketitle
\thispagestyle{empty}
\pagestyle{empty}

\begin{abstract}
Visual place recognition (VPR) is an essential component of robot navigation and localization systems that allows them to identify a place using only image data. VPR is challenging due to the significant changes in a place’s appearance driven by different daily illumination, seasonal weather variations and diverse viewpoints. Currently, no single VPR technique excels in every environmental condition, each exhibiting unique benefits and shortcomings, and therefore combining multiple techniques can achieve more reliable VPR performance. Present multi-method approaches either rely on online ground-truth information, which is often not available, or on brute-force technique combination, potentially lowering performance with high variance technique sets. Addressing these shortcomings, we propose a VPR system dubbed Multi-Sequential Information Consistency (MuSIC) which leverages sequential information to select the most cohesive technique on an online per-frame basis. For each technique in a set, MuSIC computes their respective sequential consistencies by analysing the frame-to-frame continuity of their top match candidates, which are then directly compared to select the optimal technique for the current query image. The use of sequential information to select between VPR methods results in an overall VPR performance increase across different benchmark datasets, while avoiding the need for extra ground-truth of the runtime environment.

\end{abstract}

\section{Introduction}
\label{intro}

\begin{figure}[!t]
\vspace*{1ex}
\centering
\includegraphics[width=1\columnwidth]{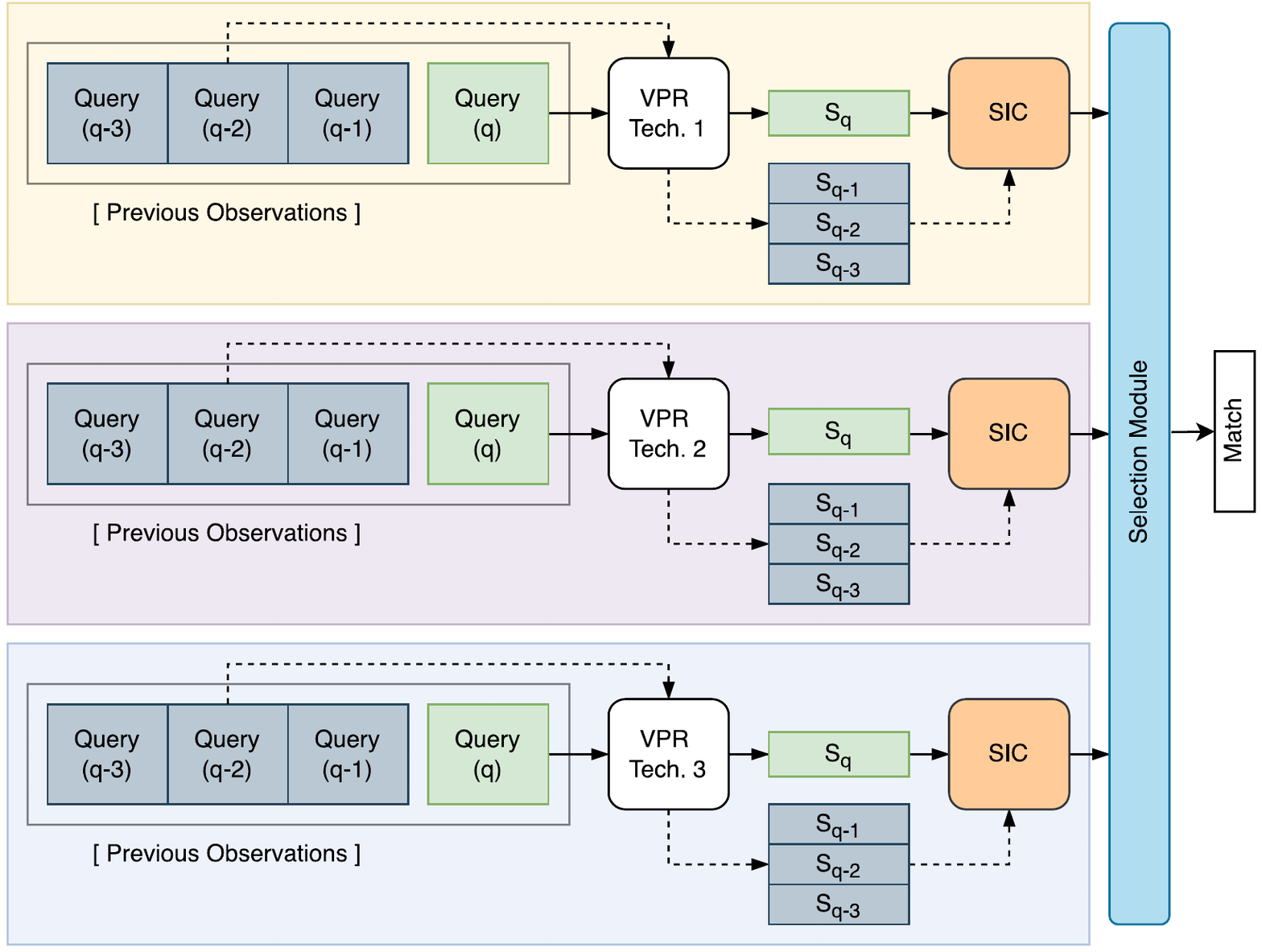}
\caption{MuSIC operating with 3 VPR techniques. For a given query image $q$, SIC analyses recent observation score vectors $S_{q-1}, S_{q-2}, S_{q-3}$ for each respective technique, outputting their sequential consistency scores. The output of SIC is then used to select which technique is used to deliver the match.}
\label{fig:VA}
\end{figure}

Visual place recognition (VPR) allows a mobile platform to localize itself in the runtime environment using image information, an attractive feature due to the low cost, versatility and availability of cameras \cite{ref:vpr-survey}. VPR is challenging due to the variety of ways in which a place's appearance can change. Changes in illumination \cite{ref:illu_changes}, seasonal variations \cite{ref:season_changes}, and varying viewpoints \cite{ref:pov_changes} can make the same place appear vastly different. Many techniques have been proposed to tackle these challenges, but no standalone technique excels in every viewing condition \cite{ref:maria_compl}.

Combining multiple techniques into a single VPR algorithm can compensate for individual weaknesses, as demonstrated by systems such as \cite{ref:mpf, switchhit}. However, \cite{ref:mpf} combines a static number of techniques for every query frame, potentially lowering the VPR performance of the optimal method when the remaining set performs poorly. \cite{switchhit} shows how switching between techniques can be a viable alternative, but it relies on additional ground-truth information of the runtime environment and technique complementarity \cite
{ref:maria_compl}.

Embedding sequential information into the VPR pipeline has also been shown to significantly improve VPR performance \cite{milford2012seqslam}. Typically, these sequence-based methods operate as an added step on top of a chosen technique. The technique pre-selection heavily influences the VPR performance of the overall system, which can potentially lead to poor VPR reliability if the method does not perform reasonably well in the target environment. 

In this work, we combine features from multi-method and sequence-based approaches to address their respective shortcomings. We propose Multi-Sequential Information Consistency (MuSIC): a VPR system which dynamically selects a technique on a per-frame basis by comparing the sequential cohesion of all techniques in the working set. The core of the approach is the Sequential Information Consistency (SIC) algorithm. SIC operates on a single technique, computing the frame-to-frame similarity continuity over recent query score vectors and outputting the most sequentially consistent reference place out of the technique's top match candidates. Observable in Fig. \ref{fig:VA}, MuSIC runs SIC on all VPR methods individually, outputting their respective best sequential consistencies for the current query, and then selects the most consistent technique to perform VPR. The proposed system improves VPR performance both by embedding sequential information and by selecting between a set of VPR techniques, while avoiding the requirement for extra ground-truth information or combining methods in a brute-force approach.

The rest of this letter is organized as follows. In Section \ref{relatedwork} we provide an overview of different approaches to VPR, with a focus on technique combination and sequence-based methods. Section \ref{method} presnts our methodology, detailing the implementation of SIC and MuSIC. In Section \ref{exp_setup} we provide the settings of our experimentation. We show and analyse our results in Section \ref{results}. In Section \ref{conclusions}, we highlight the benefits and limitations of our methods, finishing by suggesting future research paths.

\begin{figure}[!t]
\vspace*{1ex}
\centering

\includegraphics[width=0.9\columnwidth]{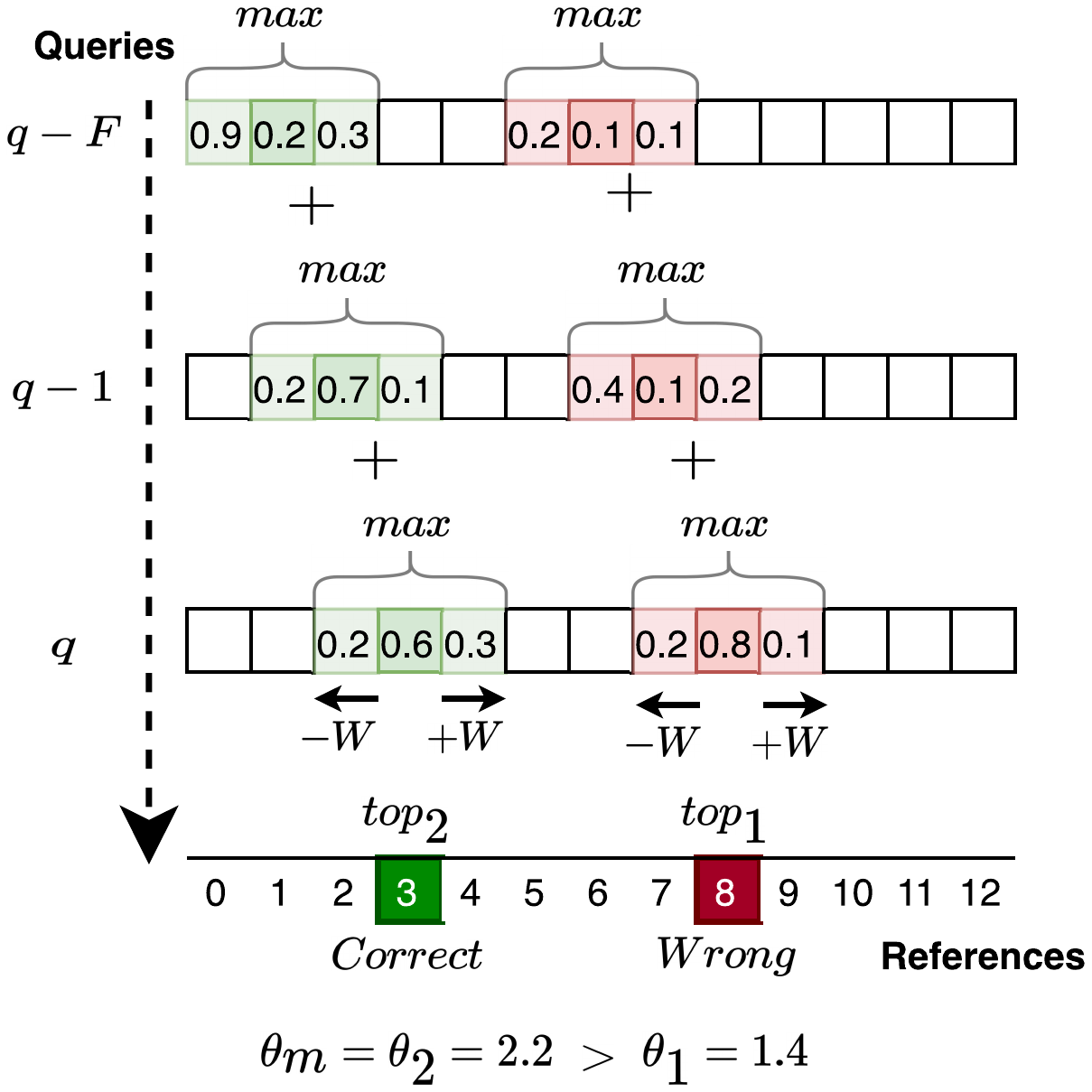}

\caption{SIC operating with $K=2, F=2, W=1$. The correct match for the current query $q$ is the reference place $3$. However, the reference place $8$ erroneously achieves the highest similarity. By computing the sequential consistency $\theta$ from the frame-to-frame continuity of previous similarity vectors, SIC identifies the correct match.}
\label{fig:sic_diagram}
\end{figure}

\section{Related Work}
\label{relatedwork}

Appearance-based localization continues being an important research topic, with several approaches being proposed in the literature. The underlying technology on which these techniques are based on varies widely. 

\cite{ref:fabmap} utilizes hand-crafted feature descriptors, such as Scale Invariant Feature Transform \cite{ref:sift} and Speeded Up Robust Features \cite{ref:surf}, to successfully build a landmark representation of the environment. While the use of local features improves resilience against viewpoint variations, it makes the method sensitive to appearance changes such as different illumination conditions \cite{cummins2008fabmap}. Well-studied global descriptors like Histogram of Oriented Gradients (HOG) \cite{ref:hog} and \cite{ref:gist} have also been employed in VPR \cite{ref:mcmanus2014scene}, but struggle with viewpoint changes. CoHOG \cite{ref:cohog} utilizes a region-of-interest approach in conjunction with HOG to minimize this shortcoming. \cite{rmac} is another popular VPR technique based on the computation of regions of interest in an image.

Image features retrieved from the inner layers of Convolutional Neural Networks (CNNs) have been shown to outperform hand-crafted features \cite{ref:cnn_for_vpr1, ref:cnn_for_vpr2}. Techniques such as HybridNet, AMOSNet \cite{ref:hybridasmosnet} and NetVLAD \cite{ref:netvlad} have utilized these CNN-extracted features to perform state-of-the-art VPR. Computational efficiency is another added consideration, with CALC \cite{ref:calc} being an example of a CNN-based VPR technique designed to perform lightweight VPR.

The variety in available image-processing methods, with different sets of strengths and weaknesses, has led to recent research on the combination of multiple techniques to perform VPR in changing environments. Multi-Process Fusion (MPF) \cite{ref:mpf} proposes a system which combines four VPR methods utilizing a Hidden Markov Model to further infuse sequential information. \cite{hier_mpf} fuses techniques in a hierarchical structure, passing only the top place candidates of upper level techniques down to the lower tiers. ROMS \cite{ROMS} (Robust Multi-Modal Sequence-based) also leverages a set of different descriptors with a sequential confirmation step to achieve higher VPR performance. The individual problem of aggregating image descriptors has been explored in \cite{HDC_agg_imgs_descs}, where the authors propose an efficient combination framework which allows for encoding of extra, possibly non-visual, information. Fusion based methods successfully improve VPR performance when the majority of the baseline techniques perform reasonably. However, fusion tends to work poorly in cases where most individual techniques do not achieve reliable VPR, potentially achieving lower performance than the best individual technique in the working set. 

SwitchHit \cite{switchhit} addresses this shortcoming by proposing an alternative multi-technique approach based on switching between techniques rather than combining them. The method starts by running a single technique and, using prior knowledge of VPR performance in the environment, decides if another technique should be run, repeating the process until a satisfactory confidence threshold is achieved. While the VPR system needs to be compliant with the worst case scenario where all techniques need to be evaluated, this approach can have the added benefit of saving power, important for mobile robotics. Nevertheless, the requirement for additional ground-truth information of the runtime environment is a major drawback, as it is often unavailable in real-world applications.

In this work, we address the shortcomings of the fusion based VPR systems without requiring any additional information of the deployment environment. Rather than relying on extra environmental knowledge, our proposed method selects between techniques by evaluating their sequential consistency. Embedding information from a sequence of place observations is a popular approach to improve VPR performance. Rather than matching a single-frame, \cite{milford2012seqslam, ref:ratslam, hoseqplacereco} compute a similarity score for entire trajectories consisting of previous observations and then take the highest scoring local candidate as the current match. \cite{ABLE-M} applies the same principle but using binary image-sequence representations, resulting in potential efficiency benefits. 

\section{Methodology}
\label{method}
  
SIC performs a search over recent queries to compute the frame-to-frame similarity continuity of the top reference candidates for the current query. MuSIC utilizes SIC on multiple techniques to select which to employ on a per-query basis. The remaining of this section details our two proposed methods.

\subsection{Problem Formulation}

Visual place recognition is often cast as an image retrieval task \cite{whereisyourplace}. A database of reference template images, commonly image descriptors, is assumed to be pre-existing, and the goal of VPR is to match the currently observed place with one of these templates. 

During navigation, a technique performs VPR for the current observed frame $q$ and a similarity vector $S_q$ is computed, where each element $S_{q, n}$ represents the similarity score associated with the $n^{th}$ reference template. In the standard image retrieval setting, the reference template which achieves the highest similarity is taken to be the correct match.

SIC performs a sequential search over recent score vectors, requiring these to be stored appropriately. The matrix $S$ is therefore constructed by stacking similarity vectors by time of observation, that is, $S_q$ is added immediately after $S_{q-1}$ (\ref{fig:matrix:A}). 

\subsection{Sequential Information Consistency (SIC)}
\label{sic_method}

\begin{figure}[!t]
	\centering
	\begin{subfigure}[b]{0.40\textwidth}
		\centering
		\includegraphics[width=\linewidth]{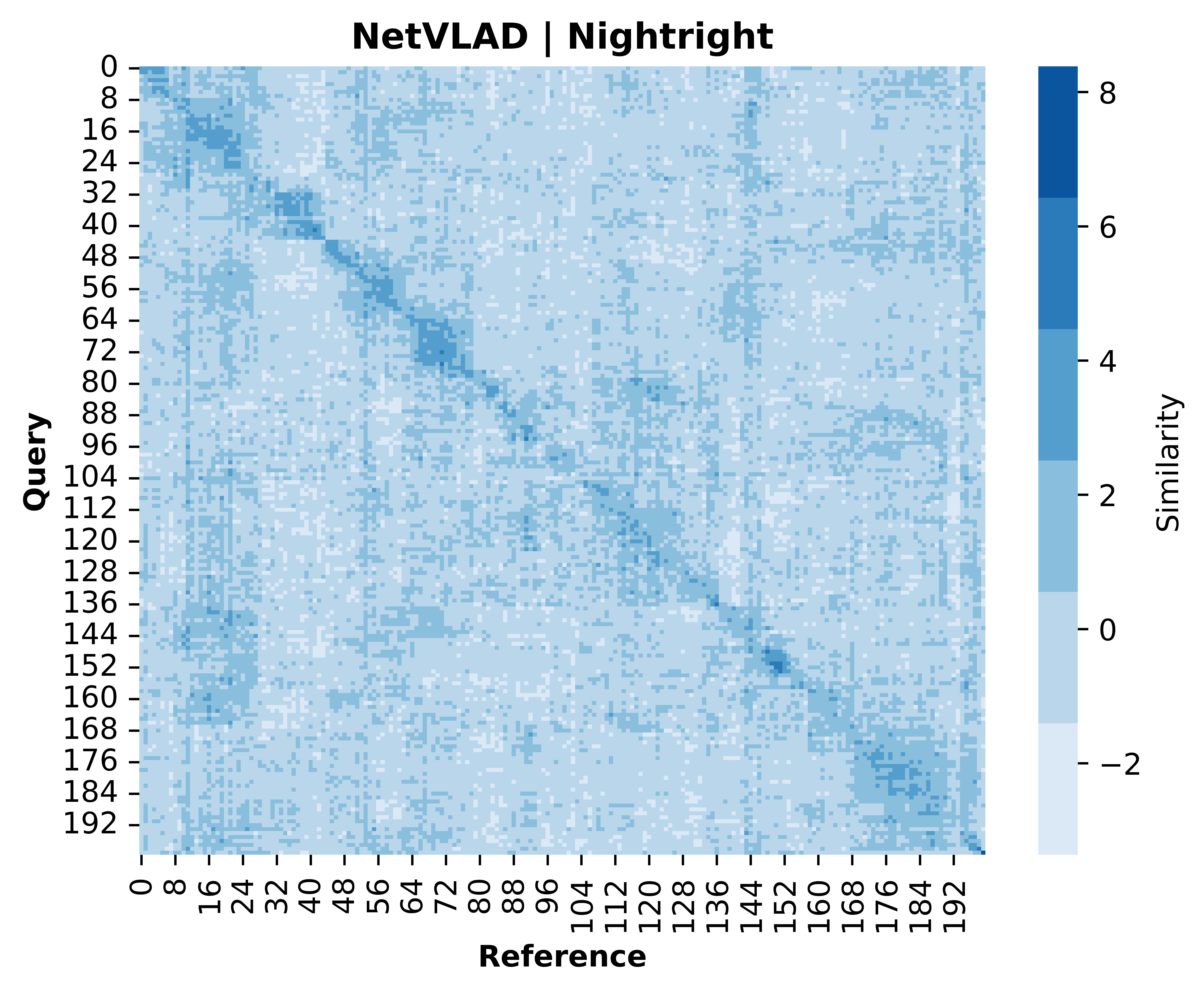}
		\caption{}
		\label{fig:matrix:A}
	\end{subfigure}
	\hfill
	\begin{subfigure}[b]{0.40\textwidth}
		\centering
		\includegraphics[width=\linewidth]{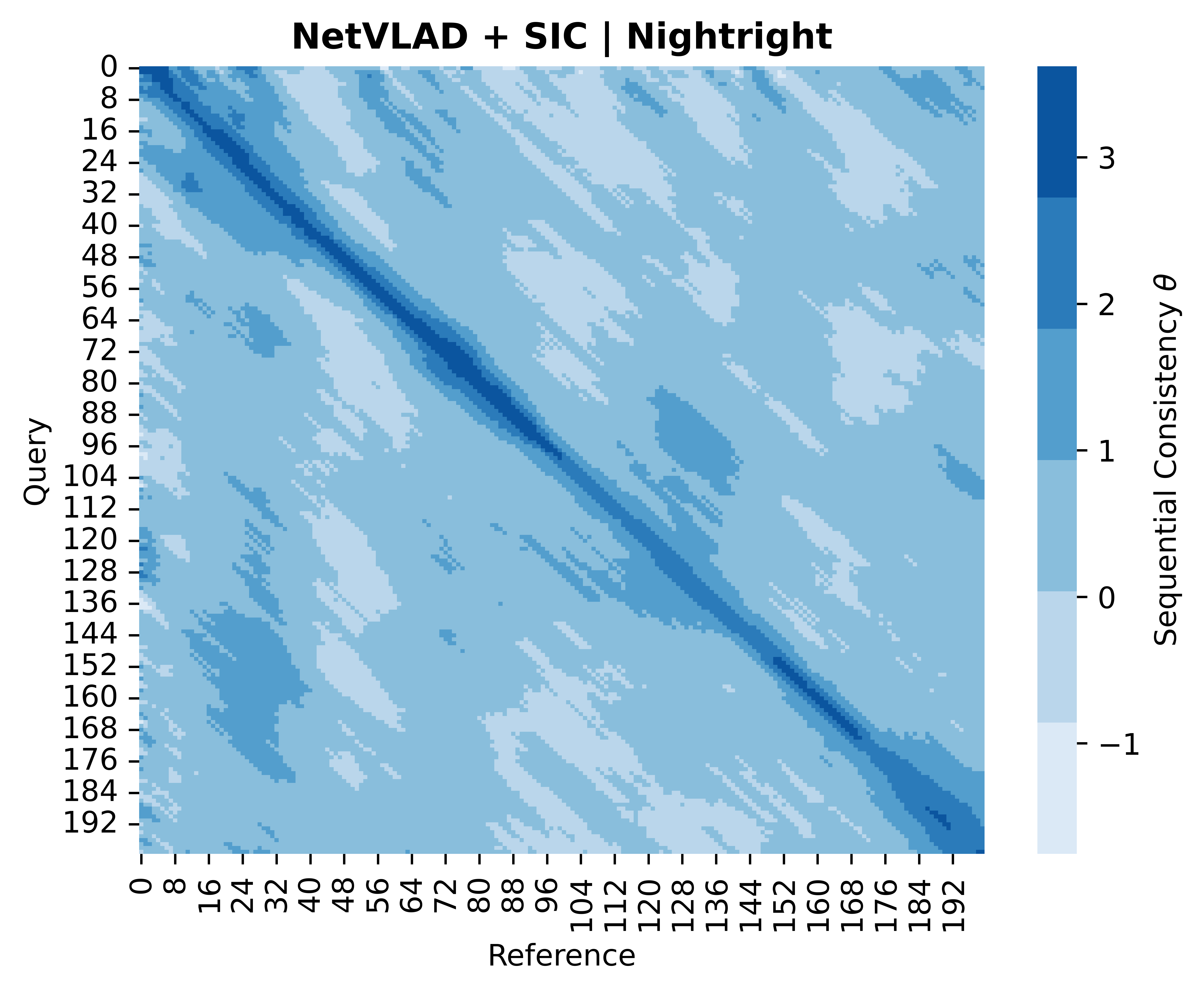}
		\caption{}
		\label{fig:matrix:B}
	\end{subfigure}

        \caption{(a) shows the usual similarity scores matrix produce, while (b) shows the sequential consistencies matrix computed by SIC (K=200, F=20, W=1)}
	\label{fig:matrices}
\end{figure}

Inspired by the trajectory similarity score introduced in \cite{milford2012seqslam} and the analysis of query candidates with highest similarity in \cite{hier_mpf}, SIC performs a search over previous query similarity vectors for the $K$ most similar templates. For each top scoring candidate $k$ in the similarity vector $S_q$ of query frame $q$, a sequential consistency $\theta$ value is calculated as follows:
\begin{gather}
  {\theta_k = \sum_{f=0}^{F} max(S_{q-f, k-f-W:k-f+W}})
\end{gather}
\noindent where $F$ denotes how many past queries should be evaluated and $k-f-W:k-f+W$ denotes a vector slice of size $W*2+1$ around the center $k-f$. Fig. \ref{fig:sic_diagram} exemplifies the computation with $K=2, F=2, W=1$. The purpose of $W$ is to relax the sequential navigation assumption and compensate for small navigation drifts. Fig. \ref{fig:matrices} shows the transformation from the regular similarity scores to the sequential consistency scores. 

The candidate which achieves the highest $\theta$, denoted as $m$, is considered to be the correct match for query $q$:
\begin{gather}
  m = \argmax_k \theta
\end{gather}
\noindent and $\theta_m$ therefore representing the maximum $theta$ value.

By evaluating only a fixed number of top matching candidates, SIC's computational cost is independent from the number of reference places in the internal map, as discussed later in Section \ref{results}. Conversely, typical sequence matching schemes suffer from increased computation times as the database of reference images increases \cite{milford2012seqslam}. Maintaining a low and stable computational profile is especially important for SIC, as it is intended to be used on multiple techniques in tandem.

\subsection{Multi-Technique SIC (MuSIC)}
\label{MuSIC_expl}


For a query image ($q$), MuSIC runs SIC with every available $t$ in $T$, generating the corresponding $\theta_m^t$ values. The candidate achieving the highest $\theta$ is then selected as the correct match. 

Different techniques have different ranges of output similarity scores. Since $\theta$ is computed directly from these values, we scale the vectors before applying SIC. Each score ${S_{q, n}}$ is normalized using the following equation:
\begin{gather}
  {\hat{S}}_{q, n} = \frac{S_{q, n} - \mu}{\sigma}
\end{gather}
\noindent where ${\hat{S}}_{q, n}$ is the scaled value, $\mu$ and $\sigma$ are the mean and standard deviation of the similarity vector ${S_q}$ currently being scaled, respectively.

SIC operates individually on each technique $t$, therefore requiring their respective $F$ past observation score vectors $S^t_{q-F:q-1}$ and current observation vector $S^t_q$. The final output match is selected by taking the reference template which achieves maximum $\theta^t_m$ amongst all techniques.

\section{Experimental Setup}
\label{exp_setup}
We conduct several experiments to evaluate the VPR performance and computational efficiency of our proposed method. This section provides details on datasets, baseline VPR techniques, technical implementations and hyperparameter settings.

\subsection{Datasets}
\label{datasets}

We conduct experiments with five benchmarks datasets: Nordland \cite{ref:nord}, Berlin \cite{berlin}, St. Lucia \cite{stlucia}, Gardens Point \cite{gardens}, and 17 Places \cite{17places}; the dataset usage is summarized in Table \ref{tab:datasets}. We allow for an error allowance of 2 frames around the ground-truth reference place for St. Lucia, 5 frames for 17 Places and 1 frame for all other datasets. If the exact frame correspondence is the reference image $c$, then an error allowance of $1$ means that reference places $c-1$, $c$, and $c+1$ are considered correct.

\newcolumntype{M}[1]{>{\centering\arraybackslash}m{#1}}
\begin{table}[t]
  \centering
  \vspace{5px}
  \caption{Dataset Details}
  \label{tab:datasets}%
    \begin{tabular}{M{1cm}M{1.5cm}M{1.5cm}M{1.5cm}M{1cm}}
    \toprule
Dataset  & Condition   & Reference Traverse  & Query Traverse & Number of Images \\
\midrule
\midrule
Nordland Winter  & Extreme seasonal    & Summer   & Winter      &   1000      \\
\midrule
Nordland Fall        & Moderate seasonal   & Summer &  Fall       &  1000   \\
\midrule
Berlin     &  Strong viewpoint & halen.-2, kudamm-1 and A100-1   &  halen.-1, 
kudamm-2 and A100-2 & 250  \\
\midrule
Night-Right & Outdoor Illumination; Lateral Shift & Day-Left     & Night-Right     &  200 \\
\midrule
St. Lucia   & Daylight; Dynamic Elements &   Afternoon   &    Morning    &   1100  \\
\midrule
17 Places  & Indoor Illumination & Day     &  Night      &    2000  \\

\bottomrule
\end{tabular}
\end{table}%

\begin{table*}[htbp]
  \centering
  \vspace{5px}
  \caption{VPR Performance: Area Under Precision-Recall Curve (AUC) and Extended Precision (EP)}
    \begin{tabular}{|c|cc|cc|cc|cc|cc|cc|cc|}
    \hline
          & \multicolumn{2}{c|}{\textbf{Winter}} & \multicolumn{2}{c|}{\textbf{Fall}} & \multicolumn{2}{c|}{\textbf{Berlin}} & \multicolumn{2}{c|}{\textbf{Night-Right}} & \multicolumn{2}{c|}{\textbf{17 Places}} & \multicolumn{2}{c|}{\textbf{St. Lucia}} & \multicolumn{2}{c|}{\textbf{Average}} \\
          \cline{2-15}
          & \multicolumn{1}{c|}{\textbf{AUC}} & \multicolumn{1}{c|}{\textbf{EP}} & \multicolumn{1}{c|}{\textbf{AUC}} & \multicolumn{1}{c|}{\textbf{EP}} & \multicolumn{1}{c|}{\textbf{AUC}} & \multicolumn{1}{c|}{\textbf{EP}} & \multicolumn{1}{c|}{\textbf{AUC}} & \multicolumn{1}{c|}{\textbf{EP}} & \multicolumn{1}{c|}{\textbf{AUC}} & \multicolumn{1}{c|}{\textbf{EP}} & \multicolumn{1}{c|}{\textbf{AUC}} & \multicolumn{1}{c|}{\textbf{EP}} & \multicolumn{1}{c|}{\textbf{AUC}} & \textbf{EP} \\
    \hline
    \hline
    HOG   & 0.28  & 0.54    & 0.85  & 0.59    & 0.03  & 0.00    & 0.03  & 0.09    & 0.18  & 0.01    & 0.70  &  0.52 &  0.35 & 0.29 \\
    
    CALC  & 0.29  & 0.51   & 0.88  & 0.5   & 0.06  & 0.01   & 0.14  & 0.02   & 0.13  & 0.01   & 0.70 &  0.51 & 0.37 & 0.26 \\
    
    CoHOG & 0.22  & 0.52   & 0.84  & 0.59   & 0.26  & 0.51   & 0.43  & 0.52   & 0.12  & 0.01   & 0.55  &  0.02  & 0.40  & 0.36 \\
    
    NetVLAD & 0.27  & 0.52   & 0.68  & 0.59   & 0.81  & 0.01   & 0.53  &  0.51  & 0.19  & 0.01   &  0.46 &  0.5  &  0.49 & 0.36 \\
    
    HOG+SeqSLAM & \textbf{0.97} & 0.6    & 0.98  & 0.88    & 0.06  & 0.56   & 0.42  & 0.71    & 0.2  & 0.03   & 0.70  &  0.51  & 0.56  & 0.55 \\
    
    CALC+SeqSLAM & 0.93  & \textbf{0.76}   & \textbf{1.00}  & \textbf{0.97}   & 0.75  & 0.56   & 0.57  & 0.52   & 0.3  & 0.03   &  0.76 & 0.52 & 0.72 & 0.56 \\
    
    CoHOG+SeqSLAM & 0.44  & 0.55  & 0.91  & 0.72   & 0.19  & 0.54   & 0.24  & 0.54   & 0.15  & 0.03  & 0.49  &  0.55  &  0.40 & 0.49 \\
    
    NetVLAD+SeqSLAM & 0.52  & 0.57   & 0.99  & 0.78   & 0.97  & 0.71   & 0.67  & 0.53   & 0.31  & 0.03   &  0.82 &  0.53  &  0.71   & 0.53 \\
    
    HOG+SIC & 0.66  & 0.51    & 0.98  & 0.82    & 0.04  & 0.03    & 0.05  & 0.50    & 0.23  & \textbf{0.51}   & 0.73  &  \textbf{0.88}  & 0.45  & 0.54 \\

    CALC+SIC & 0.92  & 0.66   & \textbf{1.00}  & 0.94   & 0.58  & 0.27   & 0.49  & 0.18   & 0.32  & 0.01   & 0.65  &  0.82  &  0.66 & 0.48 \\
    
    CoHOG+SIC & 0.72  & 0.53   & \textbf{1.00}  & 0.73   & 0.91 & \textbf{0.77}   & 0.93  & 0.68   & 0.36  & \textbf{0.51}  & 0.83  &  0.57  & 0.79 & \textbf{0.63} \\
    
    NetVLAD+SIC & 0.88  & 0.50   & 0.98  & 0.81   & \textbf{0.96}  & 0.76   & \textbf{0.98}  & \textbf{0.79}   & 0.32  &  0.01  &  \textbf{0.87}  &  0.52  & 0.83  & 0.57 \\

    MPF & 0.58  & 0.50  & 0.95  & 0.54  & 0.46  & 0.51  & 0.36  & 0.51  & \textbf{0.38}  & 0.00 &  0.43 & 0.50  & 0.53  & 0.43 \\
    
    MuSIC & 0.95  & 0.50  & \textbf{1.00}  & 0.93  & \textbf{0.96}  & 0.76  & \textbf{0.98}  & \textbf{0.79}  & 0.32  & 0.29 &  \textbf{0.87} & 0.53  & \textbf{0.85}  & \textbf{0.63} \\
  
    \hline
    \end{tabular}%
  \label{tab:vpr_perf}
\end{table*}%

\subsection{Evaluation Metrics}
\label{eval_metrics}

\subsubsection{Precision-Recall Curves}
VPR performance is often quantified using Precision-Recall curves \cite{ref:pr_just2}. The use of PR curves is favoured for class imbalanced datasets, which is the case of VPR where a small set of correct predictions exists for each query. In our setup, the Precision-Recall pairs to plot the curve are obtained by setting different confidence thresholds of the similarity between the query and the retrieved image to consider a match as correct \cite{ref:vpr_bench}. A correct VPR match is considered a True Positive (TP), an erroneous match given as correct is denoted as a False Positive (FP), and unrecognised correct matches are regarded as False Negatives (FN). Precision and Recall can then be computed by:
\begin{gather}
    Precision = \frac{TP}{TP + FP}, \,\,\,\,\,\, Recall = \frac{TP}{TP + FN}
\end{gather}
\noindent  As Precision and Recall are often in tension with one another, a wide Area Under the Curve (AUC) is indicative of good VPR performance \cite{ref:pr_jus1}, and we use this metric as an evaluation criterion.

AUC is preferred for applications needing to retrieve enough possible loop closures and incorrect matches do not result in catastrophic failure. With recent innovations in graph optimization, this setting has become the priority for VPR systems \cite{vpr_tuto}.

\subsubsection{Extended Precision}
Extended Precision (EP) \cite{EP} combines the metrics Recall at 100\% Precision ($R_{P100}$) \cite{ref:cnn_for_vpr2} and Precision at Minimal Recall ($P_{R0}$):
\begin{gather}
    EP = \frac{P_{R0} + R_{P100}}{2}
\end{gather}
Differently from AUC, EP assesses the maximum Recall value for which Precision is still at 100\% while also relaying the lower performance bound, being therefore preferred for applications where a single FP would result in severe system failure \cite{bai2018sequence}.

\subsubsection{Computation Time Per Frame}
We assess computational cost by computing the time required to match a single query, measured in milliseconds (ms). As the focus of this work is on algorithms which act as an additional step on top of baseline techniques, we exclude the underlying technique's matching time from the computation.  

\subsection{Implementation Settings}


\label{hyper_ablation}
We use the implementations available in \cite{ref:vpr_bench} for HOG, CoHOG, CALC and NetVLAD, all with the provided default settings. MuSIC is not required to operate with a specific number or combination of VPR techniques, but this set provides some variety in baseline approaches: handcrafted descriptors (HOG and CoHOG), lightweight CNN (CALC), and costly CNN (NetVLAD). To allow for fair comparison between multi-technique systems, MPF was deployed to use the same set of techniques as MuSIC, with the remaining settings as per \cite{ref:mpf}. SeqSLAM parameters are as given in \cite{milford2012seqslam}.

As detailed in Section \ref{method}, MuSIC contains three hyperparameters: $K, F,$ and $W$. We select the same parameters for all datasets with the goal of providing a general configuration. $K$, therefore, is set to 200: the number of images of the smallest benchmark dataset. For fair comparison with SeqSLAM, $F$ is set to 20: the same number of searched past observations. Finally, $W$ is set to 1, as it showed the best average performance across all datasets.

\section{Results}
\label{results}

\begin{figure}[!t]
\vspace*{1ex}
\centering
\includegraphics[width=0.85\columnwidth]{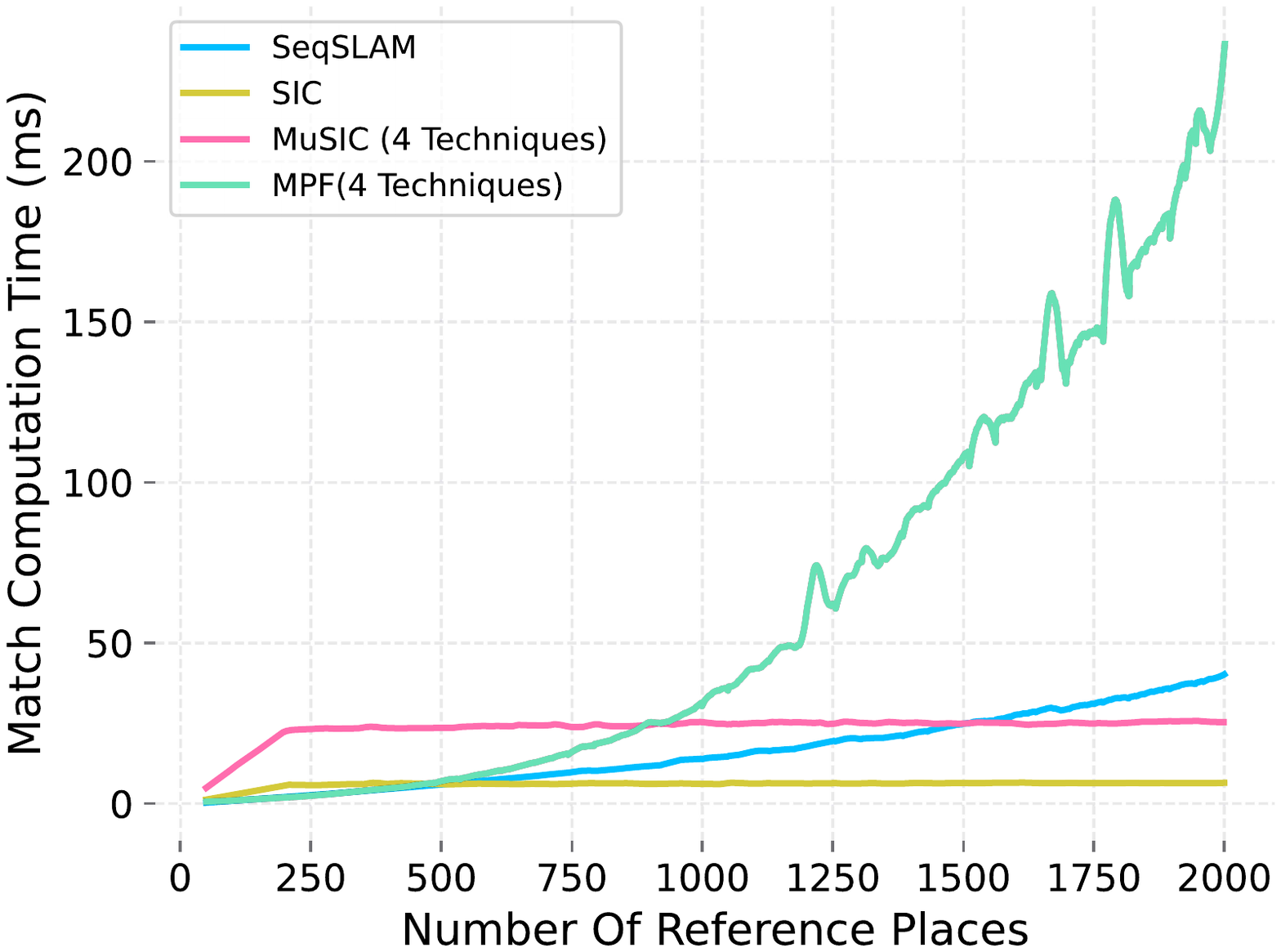}
\caption{Computational time, in milliseconds, required to match a single query frame at different map sizes, excluding baseline technique computation.}
\label{fig:comp_times}
\end{figure}

This section presents our results comparing our proposed methods with the individual baseline techniques, baselines plus embedded sequential information and multi-technique fusion. The VPR performance results are summarized in Table \ref{tab:vpr_perf}, with bold text for the best performance in each setting. Fig. \ref{fig:comp_times} displays the match computation times for increasing internal map sizes. 

\subsection{Comparison To Baselines Plus Sequential Information}
\label{res_bases_fusion}

Regarding average VPR performance, Table \ref{tab:vpr_perf} shows that, with the exception of HOG, individual techniques plus SIC achieve higher average AUC than their SeqSLAM counterparts. The best performing baseline technique with infused sequential information is NetVLAd+SIC, at 0.83 AUC, with SeqSLAM achieving its maximum when pared with CALC, at 0.72 AUC. The results are similar in terms of EP. The highest average EP for single technique plus sequential step is achieved by CoHOG+SIC, at 0.63 EP, with the highest for SeqSLAM at 0.56, again when combined with CALC.

MuSIC outperforms every individual technique with embedded sequential information, at an average AUC of 0.85 and an average EP of 0.63. With the exception of Winter and 17 Places, MuSIC achieves the same performance as the best performing individual technique plus SIC. On the Winter dataset, MuSIC achieves higher AUC than the best SIC boosted method (CALC+SIC), showing that the techniques can compensate for each other's erroneous frames even within the same dataset. On the other hand, 17 Places achieves less performance than CoHOG+SIC, indicating that the switching fails to work in this dataset. We provide a deeper analysis of the switching pattern in Section \ref{tech_sel}.

With respect to computational cost, we can observe in Fig. \ref{fig:comp_times} that, starting from internal map sizes containing more than 500 reference places, SIC computes a match faster than SeqSLAM. As explained in \ref{sic_method}, this is due to the constant number of candidates evaluation $K$. 

\subsection{Comparison To Fusion}
\label{res_sequence_methods}

The AUC results in the Night-Right dataset illustrate the downside of fusion approaches such as MPF. While the underlying techniques NetVLAD and CoHOG perform well, CALC and HOG bring the VPR performance of MPF down, even lower than the two individually best techniques. On the other hand, MuSIC correctly identifies that best performance can be achieved by running NetVLAD (\ref{fig:sels:D}). In cases where fusion does improve overall performance, such as Winter and Fall, MuSIC still achieves better VPR performance. 

Comparing the computational performance of MuSIC and MPF, the former requires less time to compute a match beginning from a map size of around 850 reference places. For larger map sizes, MPF scales significantly worse than MuSIC, which achieves a stable computational time regardless of map size.

  


\subsection{Analysing Technique Selection}
\label{tech_sel}
\begin{figure*}[!t]
	\centering
	\begin{subfigure}[b]{0.32\textwidth}
		\centering
		\includegraphics[width=\linewidth]{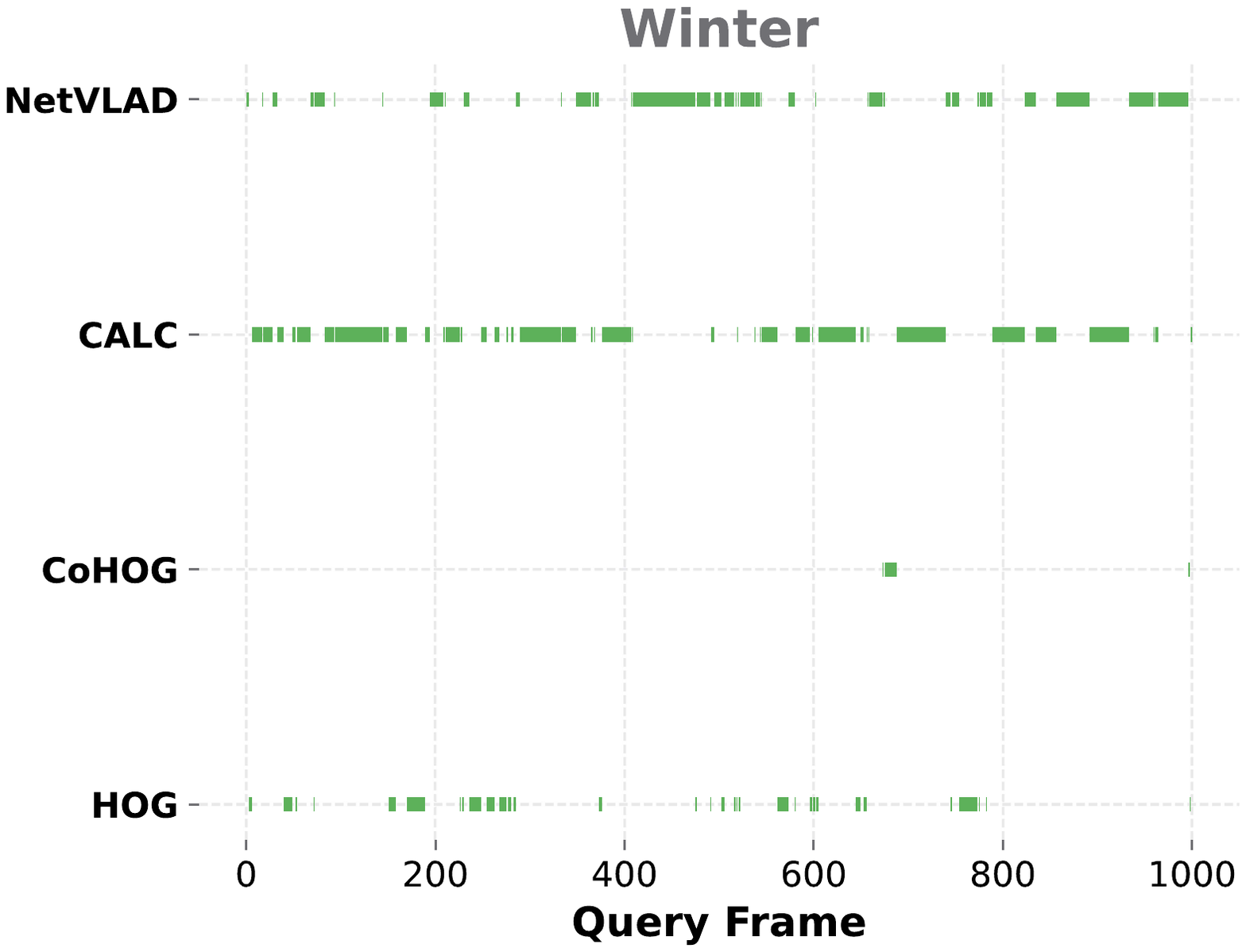}
		\caption{}
		\label{fig:sels:A}
	\end{subfigure}
	\hfill
	\begin{subfigure}[b]{0.32\textwidth}
		\centering
		\includegraphics[width=\linewidth]{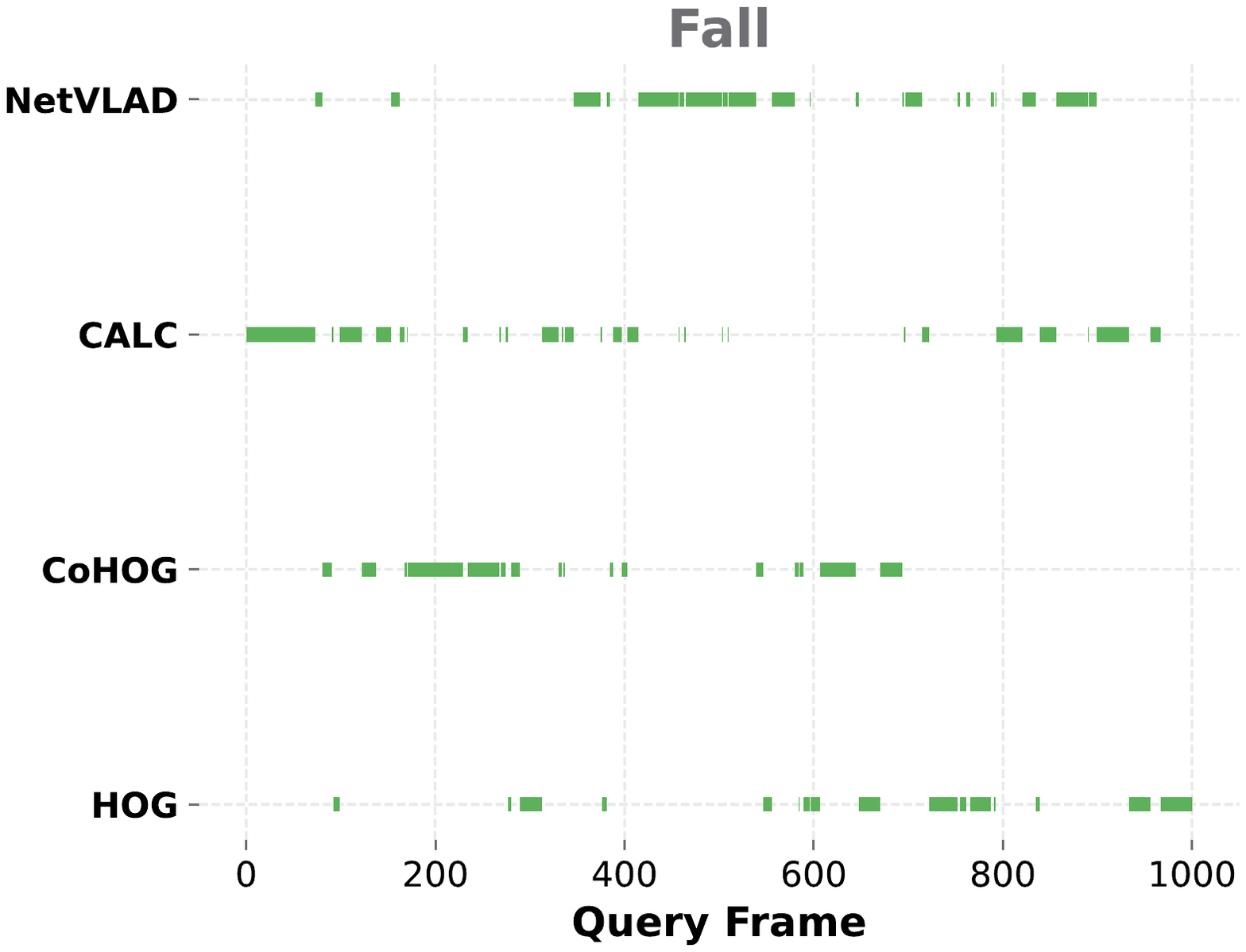}
		\caption{}
		\label{fig:sels:B}
	\end{subfigure}
	\hfill	
	\begin{subfigure}[b]{0.32\textwidth}
		\centering
		\vspace{2ex}
		\includegraphics[width=\linewidth]{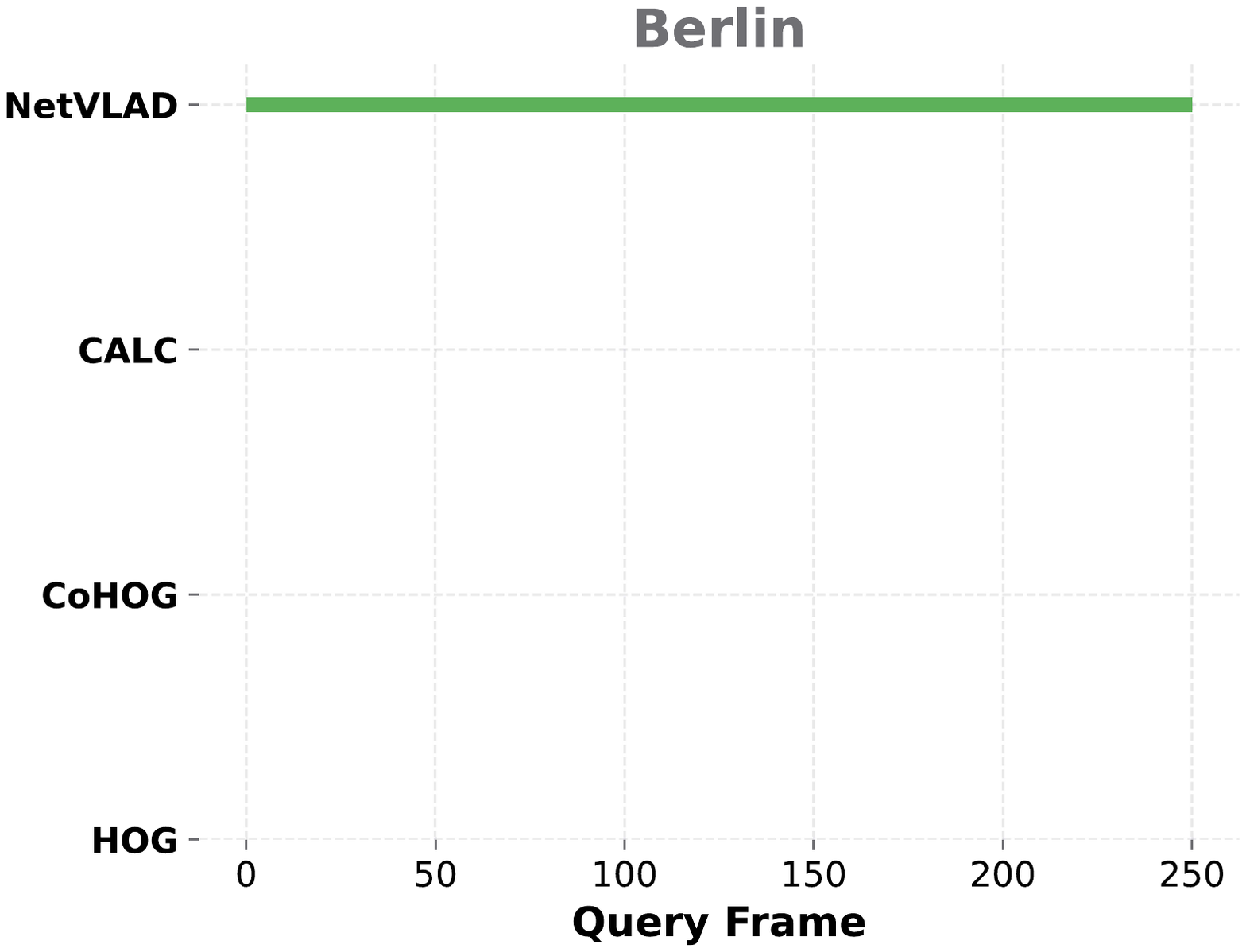}
		\caption{}
		\label{fig:sels:C}
  
	\end{subfigure}	
        \begin{subfigure}[b]{0.32\textwidth}
		\centering
		\includegraphics[width=\linewidth]{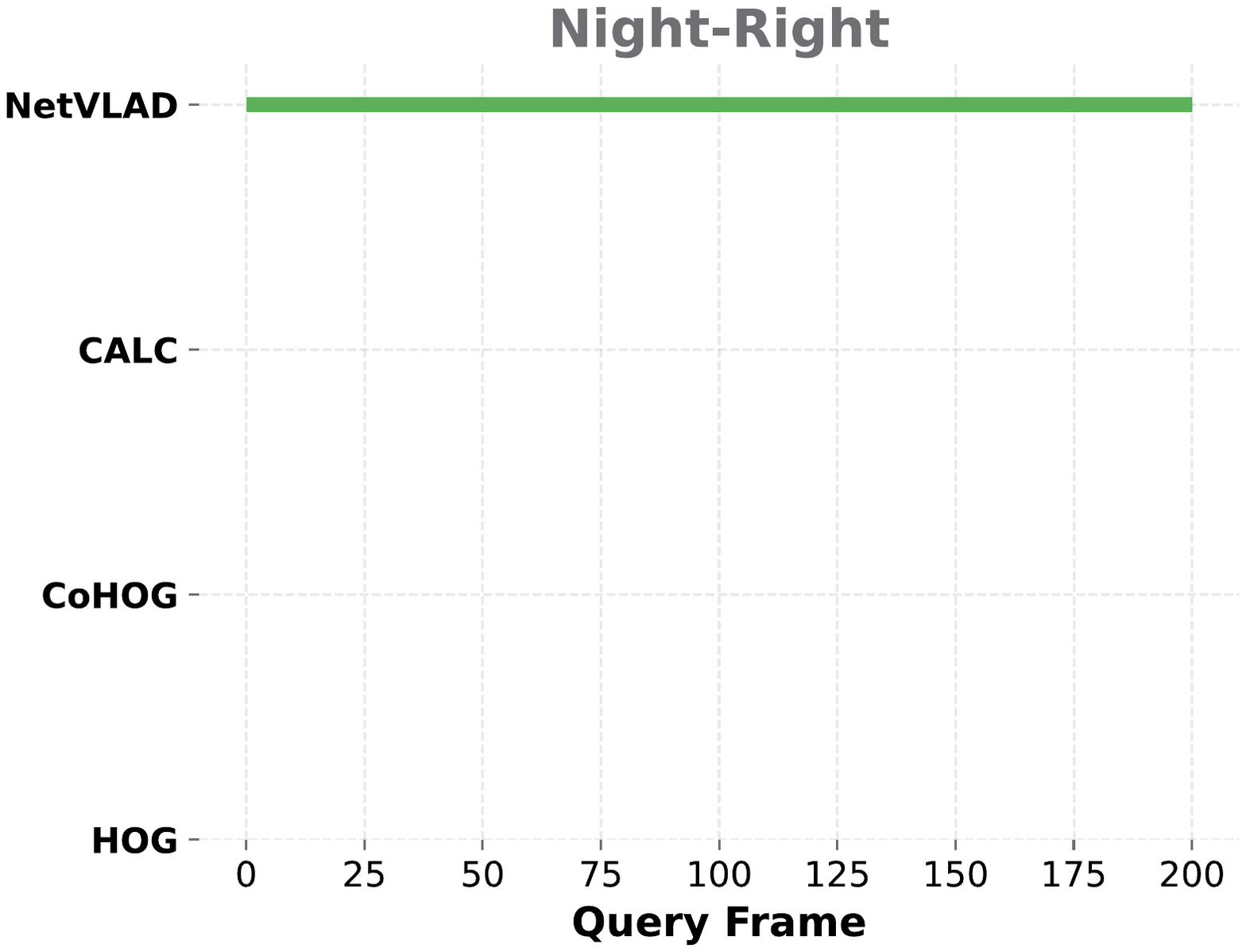}
		\caption{}
		\label{fig:sels:D}
	\end{subfigure}
	\hfill
	\begin{subfigure}[b]{0.32\textwidth}
		\centering
		\includegraphics[width=\linewidth]{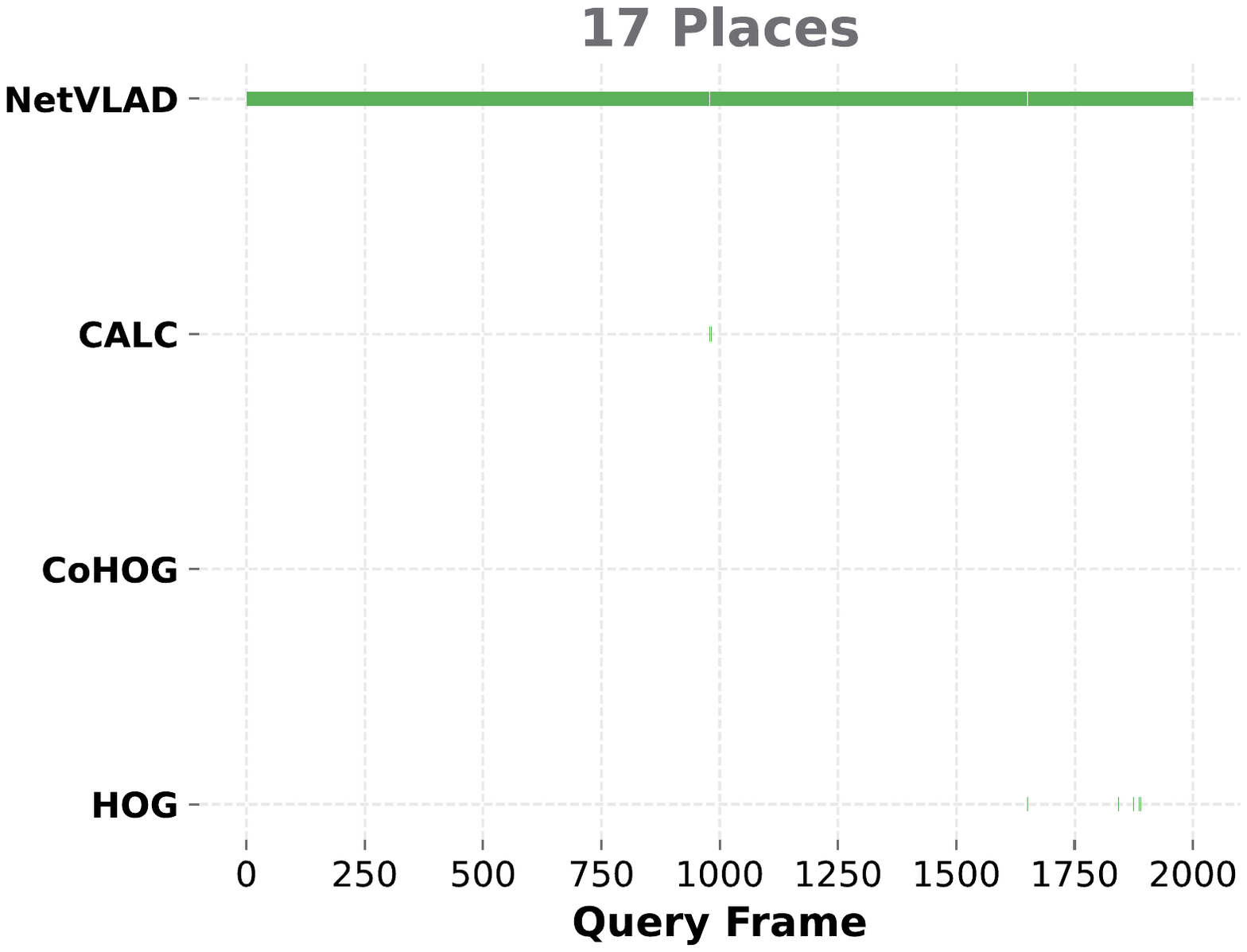}
		\caption{}
		\label{fig:sels:E}
	\end{subfigure}
	\hfill	
	\begin{subfigure}[b]{0.32\textwidth}
		\centering
		\vspace{2ex}
		\includegraphics[width=\linewidth]{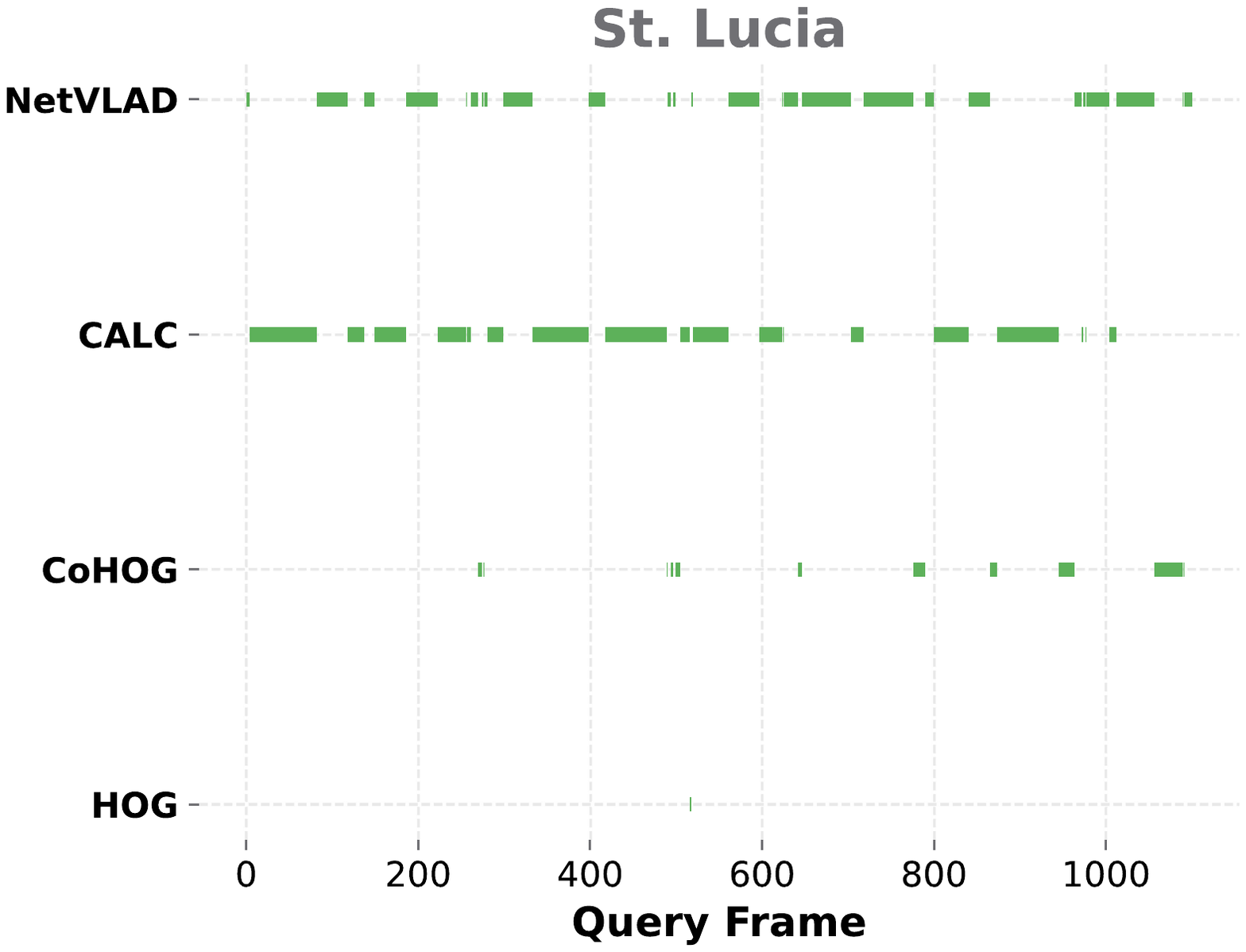}
		\caption{}
		\label{fig:sels:F}
	\end{subfigure}	
	\caption{MuSIC Technique Selections}
	\label{fig:music_selections}
\end{figure*}

The selection pattern produced by MuSIC is observable in Fig. \ref{fig:music_selections}, where we observe two distinct cases. In the plots \ref{fig:sels:C}, \ref{fig:sels:D} and \ref{fig:sels:E}, NetVLAD is the clear dominant technique, being always selected by MuSIC. In the Berlin and Night-Right datasets, this is congruent with the VPR performance metrics, as NetVLAD+SIC obtain the highest AUC and EP (Table \ref{tab:vpr_perf}). This is not consistent in the 17 Places dataset, where, according to VPR performance, CoHOG should be the dominant techniques. 

The plots \ref{fig:sels:A}, \ref{fig:sels:B} and \ref{fig:sels:F} show the second distinct case, where multiple techniques contribute to the VPR performance of the system. In terms of AUC, we observe that the lower performance bound is given by best performance technique (St. Lucia), while the highest performance bound can be higher than any of the individual techniques (Winter). According to \cite{ref:maria_compl}, the latter should occur when there is high complementarity between the underlying techniques. 

The relation between MuSIC and the individual techniques boosted by SIC is not as clear when considering EP. On each given dataset, MuSIC achieves lower EP than the respective highest individual technique plus SIC. However, on average, MuSIC achieves an EP value of 0.63, equalling that of CoHOG+SIC - the highest SIC boosted baseline. We attribute this behaviour to how SIC is designed to maximize score distribution, making the system more suitable for applications where retrieving all possible matches is prioritized over guaranteeing no false positives.

Overall, MuSIC successfully identifies the correct technique to be used per frame, allowing for a system which can be deployed on a wide range of environments without extra ground-truth information. 

\section{Conclusions, Limitations and Future Work}
\label{conclusions}
In this work, we propose a multi-technique VPR system which uses the frame-to-frame sequential continuity of several VPR techniques to select which should be used to perform VPR on the current query image. We introduce SIC, an algorithm which performs a search over recent observation score vectors produced by a single technique. SIC quantifies the sequential consistency of the top match candidates for the current query frame, resulting in a significant VPR performance improvement. MuSIC employs SIC on a set of VPR methods and, comparing their respective maximum sequential consistencies, selects which to select for performing VPR. Choosing from multiple techniques provides an additional VPR quality increase, allowing better overall performance across different datasets. MuSIC provides an alternative system to current fusion and switching methodologies, with the advantages of not requiring ground-truth information nor brute-force technique combination.

The main limitation of our proposed methodology is the sequential navigation assumption. While this assumption holds in many navigation tasks, and we have tried to atone it in our SIC algorithm, the system still heavily relies on some degree of sequentiality. Another drawback is the requirement to run every technique in the set for every query image; depending on the underlying subset of techniques, this can result in a computationally expensive VPR system. We encourage future research in addressing these shortcomings and on the general topic of online switching between VPR techniques using only data that can be computed at runtime.





\bibliographystyle{IEEEtran}
\typeout{}
\bibliography{ref}

\end{document}